\documentclass[11pt]{article} 

\usepackage[margin=1in]{geometry}
\usepackage{times}
\usepackage{amsmath,amssymb,amsthm}
\usepackage{mathtools}
\usepackage{bm}
\usepackage[inline]{enumitem}
\usepackage{booktabs,comment}
\usepackage{hyperref}
\usepackage[capitalise,noabbrev]{cleveref}
\usepackage{xcolor}
\usepackage{thmtools}
\usepackage{tikz}
\usetikzlibrary{arrows.meta, positioning, decorations.pathmorphing}
\usepackage{natbib}
\usepackage{enumitem}
\usepackage{microtype}
\usepackage{placeins}
\usepackage{algorithm}
\usepackage{algpseudocode}

\newtheorem{definition}{Definition}

\newtheorem{proposition}{Proposition}
\newtheorem{corollary}{Corollary}

\newcommand{\cP}{\mathcal{P}}

\newcommand{\cX}{\mathcal{X}}
\newcommand{\cM}{\mathcal{M}}
\newcommand{\cL}{\mathcal{L}}
\newcommand{\Width}{\mathrm{Width}}
\newcommand{\uw}{\mathrm{uw}}
\DeclareMathOperator*{\argmin}{arg\,min}
\DeclareMathOperator*{\argmax}{arg\,max}

\title{Associative Memory for Non-Stationary Environments:\\
A Self-Sizing Generalization of Hopfield Networks}

\author{Xin Li\\
  Department of Computer Science\\
  University at Albany, SUNY\\
  Albany, NY 12222, USA\\
  \texttt{xli48@albany.edu}
}

\begin{document}

\maketitle

\begin{abstract}
The Hopfield network made associative memory (AM) the model system of neural computation, but it solves the problem only
for a \emph{stationary} world: a fixed set of memories, stored once into frozen weights. Real environments are
non-stationary (e.g., memories arrive over time, drift, recur, and must be told apart from noise), where the classical
formulation fails by catastrophic interference (the palimpsest problem) and by a capacity fixed in advance. We argue
that this is not a peripheral limitation but the crux: under non-stationarity, memory and learning cease to be separate
problems, and \emph{adaptation}, rather than one-shot optimization, becomes the operative capacity. We give a fresh
formulation of the AM problem for non-stationary environments and a \emph{self-sizing} continual associative memory that
generalizes Hopfield's: it stores new memories without erasing old ones (no forgetting), re-binds drifting and recurring
memories, allocates a genuinely new memory only for true novelty, and grows
its store to the environment's \emph{intrinsic} memory demand and no further. We rigorously show that this demand is
the Urysohn width of the problem and can be estimated from data via a contrastive-similarity (CS) operator. The memory's size converges to this capacity online with no preset value and no validation search, matching an oracle capacity
search. We use experiments with synthetic datasets to show that the generalization buys self-sizing and retention under
non-stationarity, \emph{not} higher per-item recall fidelity, on which it matches strong baselines. 
\end{abstract}

\section{Introduction}
\label{sec:intro}

\paragraph{Memory and learning are two sides of one coin.} It is tempting to treat memory and learning as separate
faculties: learning acquires structure, memory stores it \citep{crowder2014principles}. That separation is an artifact of assuming a \emph{stationary}
world. Once the environment can change, the two collapse into a single problem \citep{ditzler2015learning}. To learn something new is to reshape the
substrate that holds what is already known; to remember is to resist that reshaping. Every act of writing risks erasing,
and every act of preserving forecloses learning. This coupling is the content of the stability-plasticity
dilemma~\citep{grossberg1987,mcclelland1995cls}: a system plastic enough to acquire the new is, by the same mechanism,
unstable enough to lose the old. In a stationary world the dilemma is vacuous (store once, retrieve forever) and memory
and learning can be studied apart. In a non-stationary world they are the same coin, and the central question is not how
to optimize a fixed mapping but how to \emph{adapt} an existing memory as the world moves under it.

\paragraph{Associative memory: the fruit-fly problem of neural computation.} The cleanest setting in which to study this
coupling is associative memory (AM), content-addressable storage from which a complete memory is recovered from a partial
or noisy cue. AM has long served as the \emph{Drosophila} of neural computation \cite{hertz_introduction_2018}: the minimal model organism in which
storage, retrieval, capacity, and interference can be examined whole, simple enough to analyze yet rich enough to exhibit
the phenomena that matter. (The metaphor is more than rhetorical: the fly's own mushroom-body olfactory circuit is itself
an associative memory~\citep{dasgupta2017}.) Hopfield's network~\citep{hopfield1982} made the model concrete and exact: $N$
binary units, symmetric Hebbian weights $W=\sum_\mu \xi^\mu (\xi^\mu)^\top$, an energy $E(s)=-\tfrac12 s^\top W s$ whose
minima are the stored patterns, and retrieval by descent to the nearest minimum. It is rightly canonical. Modern
continuous AMs~\citep{krotov2016,ramsauer2020} and continuous-attractor networks~\citep{burak2009} extend its
capacity and its geometry, but they inherit its defining assumption.

\paragraph{Hopfield's stationary assumption, and what it costs.} The Hopfield formulation solves AM for a \emph{closed,
stationary} world. The pattern set $\{\xi^\mu\}$ is fixed and known in advance; the weights are computed once, in a single
Hebbian batch, and then frozen; the capacity is set by $N$ before any memory arrives. Nothing in the formulation provides
for memories that appear over time, that drift, that recur, or that must be distinguished from noise. When the world is in
fact non-stationary, three failures follow, none of them incidental. \emph{(1) Catastrophic interference~\citep{mccloskey1989}.} Writing new
patterns online into a fixed substrate corrupts the old: beyond a capacity of $\approx 0.14N$ the network blacks out into
spurious states~\citep{amit1985}. Bounded-synapse remedies turn the store into a \emph{palimpsest} that forgets its oldest
memories gracefully rather than catastrophically~\citep{parisi1986,nadal1986}; but it still forgets, by construction.
\emph{(2) Fixed capacity.} A store sized in advance cannot grow to meet a demand it did not anticipate, nor shrink when
the demand is small. \emph{(3) No model of identity.} A stationary AM has no notion of \emph{when} to store: presented
with a cue, it cannot ask whether this is a known memory, a drifted version of one, or something genuinely new (the very
distinctions on which adaptation depends). The palimpsest literature \citep{parisi1986,nadal1986} shows the community has long known the stationary
assumption is the binding one; what has been missing is a formulation of the non-stationary problem that says how large the
memory should be and how it should decide what to store.

\paragraph{Adaptation over optimization.} We take Ashby's position \cite{ashby1956introduction}, and argue for the importance of adaptation to memory and learning. Much of learning theory studies
optimization: given a fixed objective or distribution, find the best fixed solution \cite{boyd2004convex}. But the environments in which
intelligence is exercised are open and non-stationary \cite{sugiyama2012machine}, where a perfectly optimized fixed solution often becomes brittle - i.e., it is
correct until the world moves and then it is wrong with no recourse. What distinguishes intelligent systems in such worlds
is not the quality of any fixed solution but the capacity to \emph{adapt} \cite{elwell2011incremental}: to absorb the new without discarding the still-useful
old, to recognize when a situation is a variant of a known one and when it is genuinely novel, and to spend representational
resources in proportion to the world's actual complexity rather than a designer's guess. On this view, generalizing AM from
the stationary to the non-stationary case is not a niche extension of an old model but the step that turns a memory
\emph{device} into an \emph{adaptive} one, which places AM at the center of a theory of adaptation rather than at the edge
of a theory of storage. 

\begin{figure}[t]
\centering
\includegraphics[width=0.7\textwidth]{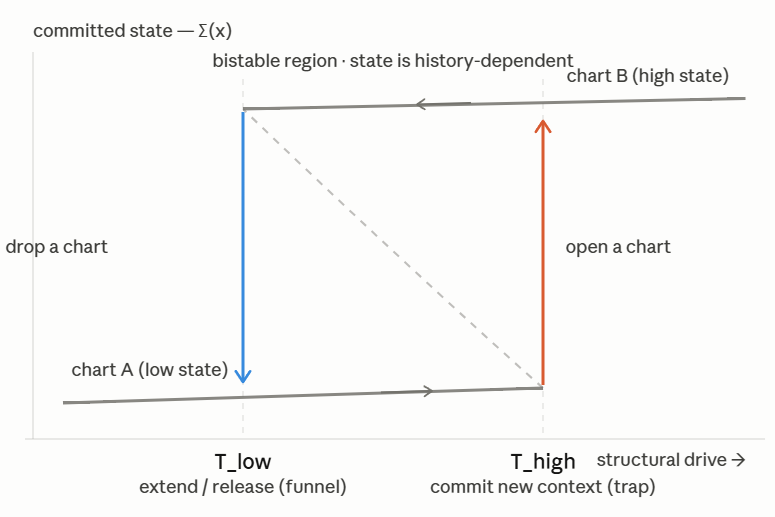}
\caption{\textbf{Structural decoupling as a hysteresis loop.} The committed memory index $\Sigma(x)$ (vertical axis) as a
function of a scalar structural drive (horizontal axis). Adaptation is held on a branch (``chart A'' or ``chart B,'' a
committed memory) by within-memory contraction (the fast, metric \emph{funnel}); the index changes only at a \emph{fold},
where the system jumps to another branch (the slow, discrete \emph{trap}). The two folds sit at different levels: committing
a new memory requires crossing the high bar $T_{\text{high}}$, while releasing the current one requires only dropping below
the low bar $T_{\text{low}}$. The gap between them is a bistable region in which the committed state is history-dependent;
it is exactly the buffer that lets the slow structural layer ignore the fast layer's moment-to-moment error. A single
threshold ($T_{\text{high}}=T_{\text{low}}$) would collapse the loop and admit boundary chatter. Hysteresis is the
dynamical fingerprint of correct temporal factorization, and present only when the problem has width $w>1$ (more than one
branch to be on); a pure funnel ($w=1$) contracts and has no loop.}
\label{fig:hysteresis}
\end{figure}

\paragraph{Adaptation factorizes in time: a slow structural layer and a fast metric one.} What makes adaptation tractable
is that its two halves run on different clocks. \emph{Within} a known memory the work is fast and continuous (contract a
noisy cue onto the stored pattern) and follow it as it drifts. \emph{Across} memories the work is slow and discrete - open a
new memory for genuine novelty, re-bind a recurrence, merge two that have become indistinguishable. These are not two
settings of one knob but two operations of different type: a \emph{metric} contraction that leaves the memory set fixed, and
a \emph{topological} change of that set \cite{nakahara_geometry_2018}. The design hazard, and the reason naive online Hopfield storage fails, is to drive
the slow discrete operation with the fast metric signal: if every fluctuation of recall error can open a memory, the store
chatters, allocating a slot per noisy cue and never settling. The remedy is a temporal factorization in which the slow layer
is held \emph{off} the fast signal and updated only on a high-bar, persistent, event-gated cue, a principle we call
\emph{structural decoupling} as shown by the hysteresis loop \cite{krasnosel2012systems} in \Cref{fig:hysteresis}. A noisy cue must clear a high bar
$T_{\text{high}}$ to commit a new memory (a structural, trap event), but only fall below a lower bar $T_{\text{low}}$ to be
released back to within-memory adaptation (a metric, funnel event); the gap between the two is a bistable, history-dependent
band that absorbs fast-signal chatter without triggering structural change. 
 
\paragraph{This paper.} We formulate the non-stationary AM problem (\Cref{sec:problem}) and give a \emph{self-sizing}
continual associative memory that generalizes Hopfield's to it. The construction stores new memories without erasing old
ones, re-binds drifting and recurring memories rather than duplicating them, allocates a new memory only for true novelty,
and grows its store to the environment's intrinsic memory demand and no further. We make ``intrinsic demand'' precise as the
Urysohn width of the problem~\citep{umtheory}, show it is estimable from data via a contrastive-separation operator
(\Cref{sec:method}), and sharpen what counts as a stable memory: not every allocation is a scaffold. We demonstrate that the memory's size converges to it online, with no preset capacity and no
validation search, and matches an oracle capacity search (\Cref{sec:experiments}). We are deliberately explicit about the
boundary of the claim (\Cref{sec:limitations}): the generalization buys \emph{self-sizing and retention} under
non-stationarity, not higher per-item recall fidelity, on which it matches strong matched-capacity baselines.

\section{From Stationary Memory to Ultrastable Adaptation}
\label{sec:problem}

We state the classical problem of AM, isolate its stationarity assumption, and then give the generalization for non-stationary setting, leading to the principle of structural decoupling.

\subsection{Problem Formulation of Associative Memory Revisited}

\paragraph{Stationary AM (Hopfield).} A stationary AM is a tuple $(\{\xi^\mu\}_{\mu=1}^P,\ \mathrm{store},\ \mathrm{recall})$.
A fixed library of $P$ patterns $\xi^\mu\in\{\pm1\}^N$ is written once, $\mathrm{store}(\{\xi^\mu\})\mapsto W$ (e.g.\ the
Hebbian outer-product rule \cite{caporale_spike_2008}), and recall maps a cue $\tilde x$ to a stored pattern, $\mathrm{recall}(\tilde x;W)=\argmin_{s}
E(s)$ under descent from $\tilde x$. Success is pattern completion: $\mathrm{recall}(\xi^\mu+\text{noise})=\xi^\mu$. The
library is fixed, $\mathrm{store}$ runs once, and $W$ never changes thereafter.
We note \emph{the hidden assumption} with the standard formulation. Three properties are baked in for Hopfield network \cite{hopfield_neural_1982}: the memory set is \emph{fixed and known} before any
retrieval; storage is a \emph{one-shot batch}; and capacity is \emph{fixed} by $N$. These are exactly the properties a
non-stationary world denies.

\paragraph{Non-stationary (continual) AM.} A non-stationary AM receives a stream $x_1,x_2,\dots$ drawn from a
\emph{time-varying} source: at each $t$ the input is a noisy realization of some latent memory $m_{c(t)}$, where the active
memory set $\{m_c\}$ may \emph{grow} (new memories appear), each $m_c$ may \emph{drift} ($m_c(t+1)$ a small perturbation of
$m_c(t)$), and memories may \emph{recur} after absence. At each step, the system must perform three coupled operations to adapt to the varying dynamics of the environment:
1) \textbf{Complete:} return the stored memory matching $x_t$ (classical recall), if one exists;
2) \textbf{Recognize:} decide whether $x_t$ is a noisy instance of a stored memory, a drifted version of one, a
  recurrence of a known memory, or genuinely novel;
3) \textbf{Update:} revise the store so that (a) old memories are \emph{not} erased by new ones (no catastrophic
  forgetting), (b) drifting and recurring memories are \emph{re-bound} to their existing slot rather than duplicated, (c) a
  new slot is allocated \emph{only} for genuine novelty, and (d) the store's size tracks the number of distinct memories the
  environment actually contains.
We call a memory satisfying (a)-(d) \emph{self-sizing}: it grows to the environment's intrinsic memory demand and no
further. 

\paragraph{What ``intrinsic demand'' is.} Desideratum~(d) requires a target: how many memories \emph{should} the store hold?
We take this to be the local Urysohn width $\uw(\cP)$ of the problem~\citep{umtheory}, the minimum number of locally
metrically-simple contexts that cover it, together with its amortized lower bound, which ties the number of simple
components needed to separate the memories to the geometry (the boundary measure, or \emph{width}) of the partition between
them. The width is a property of the environment, not of the learner. The technical core of this paper is that this target
is \emph{estimable from data} (\Cref{sec:method}) and that a simple online memory's size \emph{converges to it}
(\Cref{sec:experiments}). This is what turns the informal desideratum ``store only as much as the world demands'' into a
measurable, falsifiable claim.

\paragraph{Relation to continual learning.}
Continual learning is classically governed by the stability--plasticity
dilemma: a learner must remain stable enough to preserve previously acquired
knowledge, while remaining plastic enough to incorporate genuinely new
structure \citep{grossberg1982theory,carpenter1987art}. Desideratum~(a) alone
(no forgetting) addresses only the stability side of this dilemma and is
achievable by any method that freezes old parameters and adds new
capacity~\citep{rusu2016,yoon2018den}; by itself, it is therefore not the
central difficulty. The harder question is how much new memory should be
allocated, and when. Bayesian-nonparametric continual
learners~\citep{rasmussen2000inf,neal2000,lee2020cndpm} address this question
by inferring the number of \emph{modes} in the data. We show in
\Cref{sec:exp-amortized} that, for amortized memory, the correct target is
instead the \emph{width} of the memory partition: the number of stable memory
regions needed to preserve task-compatible structure. This width can grow even
when the apparent mode count is fixed, because a single data mode may contain
multiple incompatible memory regions that cannot be represented by one shared
amortizer without interference. The self-sizing AM we propose is, in
continual-learning terms~\citep{wang_comprehensive_2024}, an expansion method,
but its expansion is not governed by a fixed budget, a hand-tuned growth rule,
or a mode-count posterior. Instead, growth is triggered by an estimable
structural invariant. In this sense, the method resolves the
stability-plasticity tradeoff \cite{kim2023stability} by allocating plasticity only when the inferred
memory width requires a new region, while preserving stability within already
validated regions of the partition.

\subsection{Structural decoupling: holding the slow layer off the fast signal}
\label{sec:decoupling}

\paragraph{The cybernetic precedent: ultrastability.} The gap between a stationary memory and a continual one was, in fact,
identified outside the memory literature, by Ashby's analysis of \emph{ultrastability}~\citep{ashby2013design}. A stationary
Hopfield store is a single feedback system: given a cue it relaxes to the nearest minimum, a fast continuous contraction
under fixed weights. Ashby's insight was that an adaptive system facing a changing world needs a \emph{second}, slower loop
wrapped around the first - one that does not adjust the fast variables but changes the \emph{structure} (in the homeostat,
the feedback coefficients) when, and only when, the fast system fails to keep an essential variable within viable bounds. The
homeostat is therefore already the two-clock machine that non-stationary memory requires: a fast metric loop that adapts
\emph{within} a configuration, and a slow discrete loop that changes the configuration \emph{itself} \cite{kahneman_thinking_2011}. Reading the
non-stationary AM desiderata of this section against ultrastability lines them up exactly (complete and the within-memory
part of Update are the fast loop); Recognize and the structural part of Update (allocate, re-bind, merge) are the slow loop;
and self-sizing is what the slow loop achieves when it changes structure only as much as viability demands. This
correspondence is the conceptual spine of our construction, and the rest of this section develops it into a design principle, which we call ``structural decoupling''.

\paragraph{Temporally Factorized Dynamics}
The non-stationary problem asks one system to do two incompatible things at once. \emph{Within} a known memory it
must adapt continuously - i.e., contract a noisy cue onto the stored pattern, and follow that pattern as it drifts. \emph{Across}
memories it must change discrete structure - e.g., open a new memory for genuine novelty, re-bind a recurrence, merge two memories
that have become indistinguishable. The first is a metric operation that leaves the set of memories fixed (a \emph{funnel}:
contraction toward a chart's representative); the second is a topological operation that changes the set itself (a
\emph{trap}: a change of chart). The central design hazard is to drive the second with the signal that governs the first. If
every fluctuation of the fast metric error is allowed to trigger a structural change, the system chatters: it opens a memory
per noisy cue and never settles. The principle we
adopt, and that the construction of \Cref{sec:method} realizes, is \emph{structural decoupling}: the slow discrete
layer is updated only on a high-bar, persistent, event-gated signal, and otherwise deaf to the fast layer's
moment-to-moment error. 

\paragraph{The common mechanism: a fold and a gap.} The cleanest substrate for the principle is a slow-fast (singularly
perturbed) system folded over a cusp~\citep{fenichel1979geometric}: a fast variable relaxes onto a branch of a critical manifold while
a slow variable drifts along it and occasionally pushes the fast variable past a \emph{fold}, where it jumps to another
branch. In the vocabulary above, the \emph{branch identity} is the discrete chart (the trap, $c=\Sigma(x)$); \emph{relaxation
onto a branch} is within-chart contraction (the funnel, $G_c$); and the \emph{jump at a fold} is a change of chart. Such a
system exhibits hysteresis precisely when there is more than one branch to be on, so hysteresis is a dynamical witness of
width $w>1$ because a pure funnel ($w=1$) contracts and has no loop \cite{fenichel1971persistence}. The defining feature of the loop is that its two folds
sit at \emph{different} levels (Fig. \ref{fig:hysteresis}): the level at which the system commits to a new branch is higher than the level at which it
falls off the one it is on. That gap between an upper threshold $T_{\text{high}}$ (the bar to \emph{open} a new chart, a
structural event) and a lower threshold $T_{\text{low}}$ (the bar to \emph{keep extending} the current chart, a metric
admission) is the hysteresis \cite{krasnosel2012systems}, and it is exactly the mechanism that protects the slow loop from the fast loop's chatter. The
gap also interpolates between the ergodic regime (a single chart, no memory) and the strictly non-ergodic regime (folds never
crossed) \cite{walters_introduction_2000}; real systems live in the sticky middle, and the timescale separation sets where on that line they sit.

\paragraph{Ashby's ultrastability, made concrete.} The ultrastable reading (a.k.a. the
homeostat~\citep{ashby2013design}) realizes the fast/slow split mechanically. Its hysteresis lies in \emph{when}
the slow loop fires: only when an essential variable leaves its viable bounds, and once the system lands on a viable
configuration it \emph{holds} it - the configuration persists even after the disturbance that triggered the search is removed.
That persistence is memory, which is a near-literal realization of structural decoupling: the uniselector is deaf to the
moment-to-moment equilibration error of the fast loop and listens only to a thresholded, event-based novelty signal \cite{knight1996contribution}. Coupling the two, stepping the relay in proportion to fast-loop error, would
produce exactly the cross-basin chatter that decoupling is meant to prevent. Our construction generalizes ultrastability in
one respect: it not only re-wires on failure but also \emph{merges} redundant charts, which is what lets its size converge to
the width rather than remain bounded.


\paragraph{The principle of structural decoupling.} Across all three instances a single skeleton recurs: a discrete
structural variable (branch / configuration / connected component), a continuous local one (fast state / equilibrium current/ gradient magnitude), and a two-level gap that lets the structural variable update only on a high-bar, persistent,
history-dependent signal. We state the resulting design claim sharply: \emph{hysteresis is the fingerprint of correct
structural decoupling}: the loop appears precisely when a discrete chart-selection layer is being held off from the fast
signal's chatter. The construction next embodies the decoupling principle in two ways. Committed structure is
\emph{frozen}: a consolidated expert receives no gradient from later inputs, an exact decoupling (a hard gradient stop) of
what has already been committed. Structural \emph{change} (e.g., allocation, recognition, merge) is gated on the Search event (a
loss spike), but never on the within-chart contraction error, which is the approximate, hysteresis-like decoupling that governs
the plastic parts.


\section{The Urysohn Machine}
\label{sec:method}
 
Based on the argument that a non-stationary memory must hold its slow discrete layer off the fast signal, we can build the machine that the structural decoupling principle yields. We begin from the two classical architectures it generalizes because
seeing why each fails under non-stationarity is what fixes the design.
 
\paragraph{Two architectures, one shared blind spot.} The Hopfield network~\citep{hopfield1982,ramsauer2020} stores a fixed
set of patterns as attractors in a single energy landscape and recalls one by relaxing a cue downhill under frozen weights.
It is the canonical associative memory, but it is constitutively \emph{stationary}: the patterns are written once, capacity
is fixed in advance, and writing past capacity destroys recall of everything at once (the palimpsest catastrophe). It has no
notion of a memory that should arrive later, drift, or be told apart from noise. The Helmholtz machine~\citep{dayan1995helmholtz}
repairs one half of this: it is a \emph{learning} memory, with a bottom-up \emph{recognition} model that infers latent
causes and a top-down \emph{generation} model that reconstructs data, trained by the wake-sleep
algorithm~\citep{hinton1995wakesleep}. But it too assumes a \emph{fixed} latent structure (i.e., a preset number of latents fit offline to a stationary distribution) and so inherits exactly the blind spot of Hopfield network: neither architecture can
decide online that the world has changed. Hopfield fixes capacity and cannot learn the
recognition/generation pair; Helmholtz learns the pair but fixes its dimensionality. Non-stationarity demands precisely the
capability both lack: changing the number and dimension of memories as the environment reveals it.

\subsection{Metric-Topological Factorization via Structural Decoupling}

\paragraph{Factorized intelligence via structural decoupling.} The reason a single mechanism cannot tackle the non-stationarity problem is that adaptation has two parts on two clocks \cite{kahneman_thinking_2011}: a fast, continuous part that improves the response \emph{within} a known
memory, and a slow, discrete part that changes \emph{which} memories exist. Coupling them, letting the fast error move the discrete structure, makes the store chatter and re-injects the cross-memory interference that a multi-memory world forbids.
Intelligence under non-stationarity must therefore be factorized \cite{gillis2020nonnegative}: the two parts must be held on separate signals rather than driven by one. The slow part is gated on events, not on the fast error, and its decoupling takes two forms - i.e., it is exact where structure is already committed (a frozen chart receives no gradient) and approximate, event-gated, where the structure is still plastic. We call the machine
that implements this factorization the \emph{Urysohn Machine}, after Urysohn's classical bridge between metric (continuous) and topological (discrete) structure \cite{urysohn_zum_1925}. It keeps what each ancestor got right and supplies what both lacked: \emph{within} a chart, it is Hopfield (a content-addressable attractor); \emph{across} charts, it is Helmholtz (a recognition/generation pair); and
its online loop, as we will develop next, lets the structure itself grow to the environment's demand.
 
\begin{definition}[Urysohn Machine]
\label{def:um}
A \emph{Urysohn Machine} over an input space $\cX$ is a pair of coupled maps together with an update rule:
1) a \emph{recognition} map (the router) $\Sigma:\cX\to\{1,\dots,|\cM|\}$ that assigns each input to one chart of a finite
library $\cM=\{(c,G_c)\}$, and
2) a \emph{generation} family (the charts) $\{G_c\}$, each $G_c$ a locally contractive map (an exemplar/modern-Hopfield store or a local separator) that completes or predicts an input \emph{within} chart $c$;
The UM is updated online by two loops on separate signals: a \emph{fast metric loop} that, given $\Sigma(x)=c$, adapts the active
$G_c$ by the within-chart contraction error, and a \emph{slow structural loop} that adds, freezes, re-binds, or merges
charts, changing $\Sigma$ and $|\cM|$, gated only on an event (a loss spike no within-chart contraction can remove), never
on the fast error. The machine is \emph{self-sizing}: $|\cM|$ converges to the width $\uw(\cP)$ of the environment (\Cref{def:width}).
\end{definition}
 
The two loops are the two faces of one factorization: the memory factors into a \emph{metric} component (the charts $\{G_c\}$, which contract a cue within a chart and, as a by-product of the same geometry, \emph{measure} how much structure
the data demand) and a \emph{topological} component (the recognition map $\Sigma$ together with the cycle that allocates, recognizes, and merges charts, driven only by the event-gated structural signal). 
 
\paragraph{Metric component: within-chart contraction and a width estimator}

A learner receives a stream $(x_t,y_t)$ from a non-stationary environment that is a collection of \emph{contexts} (charts);
within a chart the input-output map is metrically simple (locally contractive), and charts are separated by decision
boundaries. We consider two regimes: \emph{(I)} discrete contexts arriving in blocks, and \emph{(II)} a context whose
representation \emph{drifts} continuously while occasional new contexts appear. Within a chart the metric component is a
contraction $G_c$ toward the chart's representative, an exemplar/modern-Hopfield store or a local separator, which is the
classical associative-memory operation performed inside a fixed chart.
The same geometry that makes a chart contractive also fixes \emph{how many} charts are needed. We define this count
directly.
 
\begin{definition}[Local Urysohn width]
\label{def:width}
For a problem $\cP$ on $\cX$ and a contraction modulus $\gamma<1$, the \emph{local Urysohn width} $\uw_\gamma(\cP)$ is the
minimum number of pieces in a cover of $\cX$ such that the target map is $\gamma$-contractive on each piece. When $\gamma$ is
fixed by context we write $\uw(\cP)$. Equivalently, for a classification target, it is the minimum number of locally simple
(linearly separable) charts whose union realizes the decision rule.
\end{definition}
 
Note that the Urysohn width is a property of the environment, not of any learner (like VC dimension \cite{vapnik2013nature}), and it lower-bounds the capacity any covering memory must
hold. The bound is governed by the \emph{measure of the decision boundary} $\Width_\partial$ (e.g., the surface area, in input
space, of the region where the target switches charts) because each simple chart can absorb only a bounded amount of that
boundary.
 
\begin{proposition}[Amortized capacity bound]
\label{prop:amortized}
Let a separator have decision boundary $B_\tau$ of measure $\Width_\partial(\tau)$ at scale $\tau$. Assume that each
$L$-Lipschitz basis chart that resolves the target to accuracy $\varepsilon$ can cover at most $C\varepsilon$ units of
$B_\tau$, where $C=C(d,L)$ depends only on the ambient dimension and the Lipschitz constant. If $k$ charts realize the
separator to accuracy $\varepsilon$, then
$k \;\ge\; \frac{\Width_\partial(\tau)}{C\,\varepsilon}$.
Consequently $\uw(\cP)\ge \Width_\partial(\tau)/(C\varepsilon)$.
\end{proposition}
 
\noindent
To turn the lower bound into an operational target, we estimate
$\Width_\partial$ from data. Given labeled samples
$\{(x_i,y_i)\}_{i=1}^n$ and bandwidth $h$, let
$W_{ij}=\exp(-\|x_i-x_j\|^2/h^2)$. The contrastive-similarity
(CS) operator is the within-class Laplacian
$\cL_h=D-W^{\mathrm{cs}}$, with
$W^{\mathrm{cs}}_{ij}=W_{ij}\mathbf 1[y_i=y_j]$. The
multiplicity of its zero eigenvalue estimates the number of class-connected
charts, while the normalized cross-label cut
$\widehat{\Width}_{\partial,h}
=
\frac{1}{c\,n\,h}\sum_{i,j} W_{ij}\mathbf 1[y_i\neq y_j]$
estimates boundary measure by counting kernel-weighted edges crossing the
empirical decision boundary. Here $n$ is the sample size and $c$ is a fixed
geometric normalizer, absorbing the unit-bandwidth kernel-slab constant in the
ambient dimension, so that $\widehat{\Width}_{\partial,h}$ is an intensive
estimate of boundary width rather than a quantity that grows with batch size or
bandwidth.

The width estimator has the locality needed in a streaming, non-stationary setting. Within-class blocks of $W^{\mathrm{cs}}$ are the basins of the current charts: a sample far from all committed exemplars joins no block and increases the component count, signaling demand for an additional chart. Dually, the cut charges only cross-label edges near an empirical boundary. A noisy cue within an existing chart contributes little because its neighbors share its label, whereas a cue that straddles a genuinely new decision surface produces a spike of cross-label weight. The CS estimator separates the two events a non-stationary memory must distinguish: within-chart drift, which the active contraction should absorb, and across-chart novelty, which should trigger structural change. The relevant boundary estimator is the cut above, not the spectral gap: $\lambda_1(\cL_h)$ measures bottleneck difficulty rather than boundary measure. This gives the target but not yet the online mechanism. In a batch, the cut estimates the width $\uw(\cP)$ demanded by the environment; in a stream, the same boundary event appears one sample at a time as a loss spike that no within-chart contraction can remove. The topological component turns this pointwise novelty signal into a control cycle. Ordinary fast error keeps the active chart plastic, but an irreducible loss spike gates a discrete structural move: recognize an existing chart, allocate a new one, freeze a committed expert, or merge redundant charts. The transition from the metric component to the topological component is the transition from measuring the required width to dynamically realizing it without erasing committed memory.


\begin{figure}[t]
\centering
\begin{tikzpicture}[
  phase/.style={draw, very thick, rounded corners=8pt, minimum width=2.6cm,
    minimum height=1.3cm, font=\small, align=center, fill=#1!12},
  arr/.style={-{Stealth[length=6pt]}, very thick, #1},
  lbl/.style={font=\tiny, text=gray!70!black, align=center}
]
\node[phase=green!60!black] (E) at (90:3.0)
  {\textbf{Navigate}\\[-1pt]\footnotesize Evaluate};
\node[phase=orange] (D) at (210:3.0)
  {\textbf{Search}\\[-1pt]\footnotesize Detect};
\node[phase=blue] (T) at (330:3.0)
  {\textbf{Closure}\\[-1pt]\footnotesize Transform};
\draw[arr=red!70!black] (E.240) to[bend right=15]
  node[lbl, left=3pt] {loss spike\\$\ell_t \gg \ell_{\mathrm{prev}}$} (D.80);
\draw[arr=orange!70!black] (D.350) to[bend right=15]
  node[lbl, below=3pt] {recognize\\or allocate} (T.190);
\draw[arr=green!50!black] (T.100) to[bend right=15]
  node[lbl, right=3pt] {resume with\\updated library} (E.320);
\node[font=\scriptsize\itshape, text=green!50!black] at (90:4.3) {apply quotient};
\node[font=\scriptsize\itshape, text=orange!80!black] at (220:4.3) {detect inadequacy};
\node[font=\scriptsize\itshape, text=blue!70!black] at (320:4.3) {commit quotient};
\node[draw, dashed, rounded corners=4pt, fill=gray!5, font=\scriptsize,
  align=center] at (0,-0.2)
  {At $|\cM| = \uw(\cP)$:\\Search silent,\\only Navigate};
\end{tikzpicture}
\caption{The E-D-T / Navigation-Search-Closure cycle. \textbf{Navigate} (Evaluate) applies the current separator, the
steady-state mode, during which the within-chart metric learner trains. \textbf{Search} (Detect) fires on a loss spike,
signalling structural rather than parametric inadequacy. \textbf{Closure} (Transform) commits the current structure
(freezing) and updates the library (recognition or allocation), after which a periodic merge consolidates redundant charts.
At convergence ($|\cM|=\uw(\cP)$) the cycle degenerates to pure Navigation. Outer labels give the topological operation at
each phase; inner labels give the algorithmic procedure.}
\label{fig:edt-cycle}
\end{figure}

\subsection{Topological Component: E-D-T cycle as a relaxation oscillation}
\label{subsec:oscillation}

The online dynamics is a three-phase cycle (\Cref{fig:edt-cycle}) with two equivalent readings. At the \emph{algorithmic}
level it is Evaluate-Detect-Transform (E-D-T): \textbf{Evaluate} the active separator on the current input and read its
loss; \textbf{Detect} whether that loss signals a \emph{structural} inadequacy (a shift no gradient step on the current chart
can fix) rather than ordinary parametric error; \textbf{Transform} the structure by freezing the active expert, allocating or
recognizing a chart, and updating routing. At the \emph{computational} level the same phases are Navigation-Search-Closure:
the system is exploiting a known map (Navigation), exploring because the current map has a gap (Search), or committing newly
discovered structure (Closure). Topologically each phase acts on a quotient of the input space: Navigate \emph{applies} the
current quotient $\pi_c\circ G_c\circ\phi$, Search \emph{detects} that it is too coarse, and Closure \emph{commits} a refined
quotient, freezing makes the equivalence classes irreversible and allocation splits one class in two. To this split-and-freeze
cycle we add a \emph{Merge} (consolidation) step that fuses two charts whose centroids have become indistinguishable; merge
coarsens an over-refined quotient.
We derive the cycle as forced by the following chain of reasoning:
\[
\text{non-stationarity }(\uw>1)
\;\rightarrow\;
\text{factorization}
\;\xrightarrow{\text{decoupling}}\;
\text{hysteresis}
\;\rightarrow\;
\text{E--D--T relaxation oscillation}.
\]

\paragraph{(1) Non-stationarity forces factorization.} Call a chart \emph{feasible} on regime $c$ if some $G_c$ is
$\gamma_c$-contractive there with $\gamma_c<1$ and low risk; the local Urysohn width $\uw(\cP)$ is the minimum number of
feasible charts covering the stream. Suppose a single map $G$ were $\gamma<1$-contractive across two regimes with distinct
targets $y_1\neq y_2$. By the Banach fixed-point theorem \cite{granas_fixed_2003}, a contraction on a connected component has a \emph{unique} fixed
point, so either the two regimes lie in one component (contradicting $y_1\neq y_2$) or they are metrically separated and a
discrete index $\Sigma$ is needed to distinguish them before any contraction applies. Either way the optimal map factorizes,
$f(x)=G_{\Sigma(x)}\circ\pi_{\Sigma(x)}\circ\phi(x)$, into a discrete \emph{trap} $\Sigma$ and continuous \emph{funnels}
$\{G_c\}$. Therefore, $\uw>1$, genuine regime structure, is what makes factorization mandatory.

\paragraph{(2) Decoupled factorization yields hysteresis.}
As argued in
\Cref{sec:decoupling}, the slow library must also be insulated from the fast
within-chart loss: if ordinary metric error were allowed to update the
discrete structure, the model would reintroduce the cross-regime mixing that
$\uw>1$ rules out. 
The online learning rule is implemented by allowing structural updates only when a statistic $s_t$ indicates
irreducible mismatch. Because $s_t$ is estimated from the same noisy stream that
drives fast adaptation, suppose
$|\widehat{s}_t-s_t|\le \varepsilon$. A single threshold $\tau$ is often unstable
near the decision boundary: whenever $s_t\approx \tau$, noise can push
$\widehat{s}_t$ back and forth across the threshold, causing the library to
chatter. Hysteresis removes this ambiguity by replacing one threshold with two.
A chart is opened only when $\widehat{s}_t$ exceeds
$T_{\text{high}}$, and released only when it falls below
$T_{\text{low}}$, leaving a dead zone
$\Delta\tau=T_{\text{high}}-T_{\text{low}}$. No single noise excursion can cross
both thresholds provided
$\Delta\tau>2\varepsilon$, which is the hysteresis-switching condition~\citep{morse2002applications}. 

\paragraph{(3) Hysteresis plus a slow drive is a relaxation oscillator.} Let $s$ be the structural drive and let the discrete branch be the committed chart. On a committed chart
the system runs Navigate and drains mismatch; as the non-stationary environment shifts, the active chart progressively
misfits and $s$ drifts up, $\dot s=\alpha>0$, with $\alpha$ the environment's rate of structural change. When $s$ reaches
$T_{\text{high}}$, Search detects and Closure jumps to a new (or re-bound) chart, which fits, resetting $s$ to
$s_{\mathrm{reset}}<T_{\text{low}}$. The drive is a sawtooth (slow ramp, fast reset) i.e.\ a relaxation
(integrate-and-fire) oscillator whose limit cycle \emph{is} E-D-T: the ramp is Navigate (E), the crossing of
$T_{\text{high}}$ is Search (D), the reset is Closure (T). The period self-tunes to the drive,
$T_{\mathrm{period}}\approx(T_{\text{high}}-s_{\mathrm{reset}})/\alpha$, and the dead zone guarantees a clean, well-separated
oscillation rather than chatter (the same $\Delta\tau>2\varepsilon$). 

\begin{algorithm}[t]
\caption{Hysteretic E-D-T cycle: the double-threshold gate.}
\label{alg:ttt-hyst}
\begin{algorithmic}[1]
\Require library $\cM=\{(c_j,G_j)\}$; stream $(x_t,y_t)$; recognition threshold $\theta_{\mathrm{rec}}$;
  \textbf{loss-spike dead zone} $\tau_{\mathrm{lo}}<\tau_{\mathrm{hi}}$; merge threshold $\theta_{\mathrm{mrg}}$; rate $\alpha$
\State init: allocate first expert; $a\gets1$; $\ell_{\mathrm{prev}}\gets\infty$; \textbf{$\mathrm{mode}\gets\textsc{stable}$}
\For{each input $(x_t,y_t)$ in the stream}
  \State \textbf{Navigate (Evaluate):} $a\gets\argmax_j\mathrm{sim}(x_t,c_j)$;\ $\hat y_t\gets G_a(x_t)$;\ $\ell_t\gets\cL(\hat y_t,y_t)$;\ $\rho_t\gets\ell_t/\ell_{\mathrm{prev}}$
  \If{$\mathrm{mode}=\textsc{stable}$ \textbf{and} $\rho_t>\tau_{\mathrm{hi}}$} \Comment{\textbf{rising edge clears HIGH bar $\Rightarrow$ trap}}
    \State $\mathrm{mode}\gets\textsc{committing}$
    \If{$\exists\, j:\ \mathrm{sim}(x_t,c_j)>\theta_{\mathrm{rec}}$}
      \State \textbf{Recognize:} $a\gets j$; update $(c_a,G_a)$ \Comment{re-bind: drift or recurrence}
    \Else
      \State \textbf{Allocate:} freeze $G_a$; append $(c_{\mathrm{new}},G_{\mathrm{new}})$; $a\gets\mathrm{new}$ \Comment{commit + extend partition}
    \EndIf
    \State $\ell_{\mathrm{prev}}\gets\ell_t$ \Comment{\textbf{re-anchor baseline to the new active chart}}
  \ElsIf{$\mathrm{mode}=\textsc{committing}$ \textbf{and} $\rho_t<\tau_{\mathrm{lo}}$} \Comment{\textbf{falling edge below LOW bar $\Rightarrow$ funnel}}
    \State $\mathrm{mode}\gets\textsc{stable}$
  \EndIf
  \Statex \quad\ \ \textbf{else} $\rho_t\in[\tau_{\mathrm{lo}},\tau_{\mathrm{hi}}]$: \textbf{dead zone---latch held, no structural action}
  \State \textbf{Adapt:} update active expert $G_a$ on $(x_t,y_t)$ \Comment{within-chart contraction (new or current chart)}
  \If{$\mathrm{mode}=\textsc{stable}$} $\ell_{\mathrm{prev}}\gets(1{-}\alpha)\ell_{\mathrm{prev}}+\alpha\ell_t$ \Comment{\textbf{baseline tracks on-branch only}}
  \EndIf
  \State \textbf{Merge (consolidate):} if some $i\neq j$ have $\mathrm{sim}(c_i,c_j)>\theta_{\mathrm{mrg}}$, fuse charts $i,j$
\EndFor
\Statex \textbf{At convergence} ($|\cM|=\uw(\cP)$): the drive stays in the dead zone, $\mathrm{mode}=\textsc{stable}$; Search is permanently silent, only Navigate runs.
\end{algorithmic}
\end{algorithm}

\Cref{alg:ttt-hyst} gives the operational form of the cycle. It specifies what the
learner does at each sample: route, test for structural mismatch, update the
active expert if the mismatch is ordinary, and otherwise change the discrete
library by recognition, allocation, freezing, or merge. The remaining question
is why this control law is stable rather than an ad hoc trigger rule. The
answer is that E-D-T is the discrete trace of a relaxation oscillator. Let
\(s_t\) denote the slow structural drive accumulated by mismatch between the
current chart and the stream. During Navigate, ordinary error is drained by
within-chart contraction; under non-stationarity, however, irreducible mismatch
accumulates until it crosses a high threshold. Search then fires, and Closure
performs a fast structural reset by re-binding or allocating a chart, after
which the drive falls below the release threshold. The same event gate
that prevents fast metric error from rewriting the library also produces the
slow-ramp/fast-reset dynamics of a relaxation oscillator. We summarize our findings into the following proposition and two corollaries.

\begin{proposition}[E-D-T as a relaxation oscillation of the agent-environment system]
\label{prop:oscillation}
Assume (1) $\uw(\cP)>1$; (2) the hysteresis gap satisfies $\Delta\tau>2\varepsilon$, where $\varepsilon$ bounds the
per-step estimation error of the structural statistic; and (3) Closure is effective, i.e.\ each accepted jump resets the
drive to $s_{\mathrm{reset}}$ with $s_{\mathrm{reset}}+\varepsilon<T_{\text{low}}$. Then the coupled agent-environment
dynamics execute the E-D-T cycle as a relaxation oscillation, with slow Navigate ramps punctuated by fast Closure resets and
no boundary chatter.
\end{proposition}
 
\begin{corollary}[Bounded environment: a terminating transient]
\label{cor:terminating}
If the environment has bounded width and no persistent drift, then once $|\cM|=\uw(\cP)$ the drive can no longer reach
$T_{\text{high}}$ ($\alpha\to 0$), and the oscillation halts: Search is permanently silent and only Navigate runs. Equivalently,
the Lyapunov function $V(\Pi)=a\,I_{\mathrm{tot}}(\Pi)+b\,|\cM|$ strictly decreases on every accepted nonredundant move, so the
cycle is a dissipative transient with at most $\uw(\cP)$ novel allocations rather than a perpetual limit cycle.
\end{corollary}
 
\begin{corollary}[Open-ended environment: a sustained limit cycle]
\label{cor:sustained}
If non-stationarity is sustained ($\alpha>0$ persistently - e.g., continual drift, recurrence under forgetting, or an unbounded
context stream), the same dynamics form a sustained relaxation oscillation whose frequency tracks $\alpha$. The drive that
powers the oscillation is the very non-stationarity that forced the factorization in step~(1): the environment both
necessitates the structure and energizes its dynamics.
\end{corollary}

\paragraph{The E-D-T cycle generalizes wake-sleep}
 
\begin{figure}[t]
\centering
\includegraphics[width=0.8\textwidth]{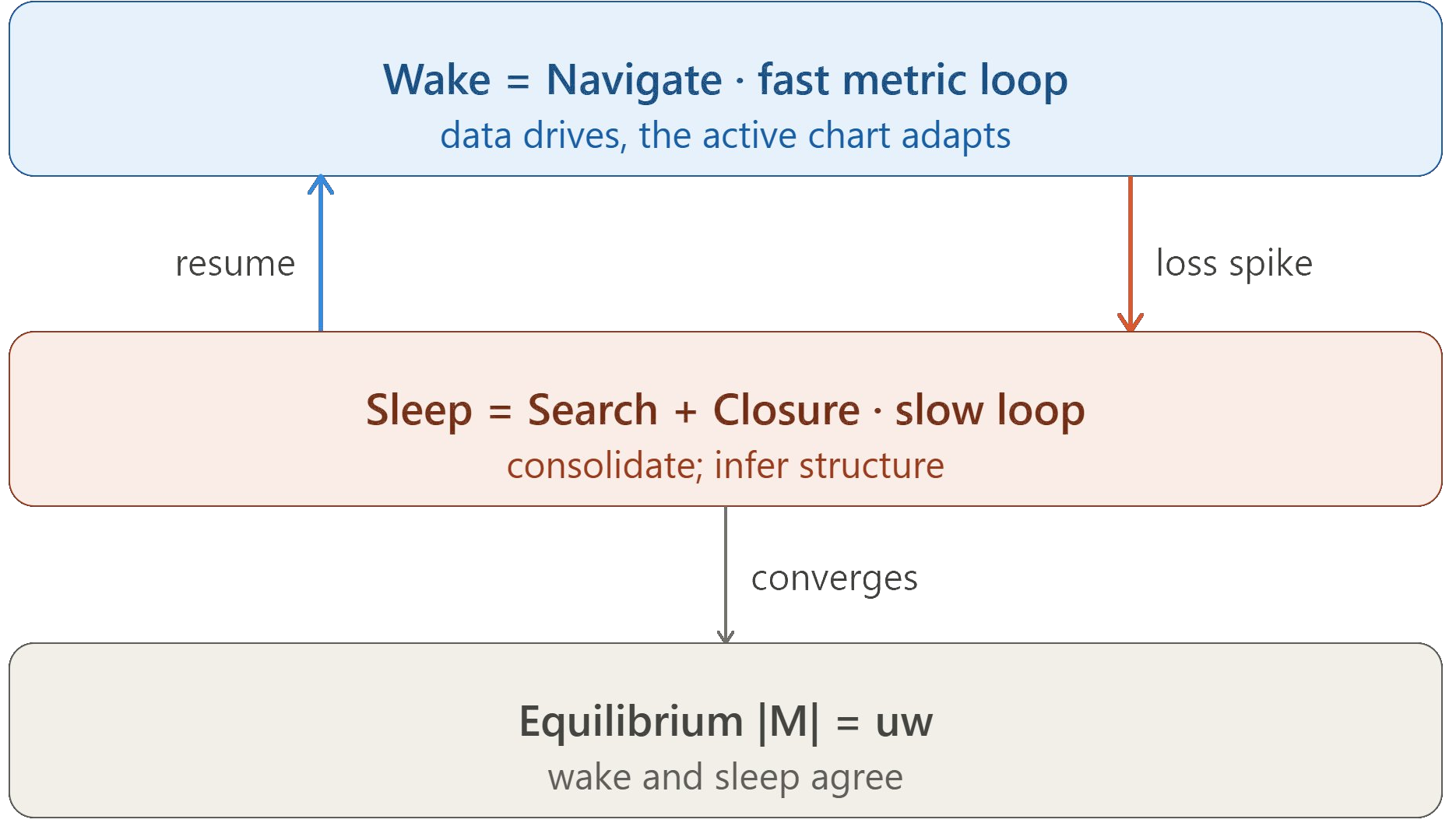}
\caption{\textbf{The E-D-T cycle as a structure-inferring wake-sleep algorithm.} Navigate is the wake phase (the fast
metric loop: data drive recognition, the active chart adapts); Search and Closure are the sleep phase (the slow structural
loop: consolidate and infer structure). Unlike classical wake-sleep, the
Urysohn Machine's sleep phase changes the number of charts; the two phases reach equilibrium exactly when the library size
equals the Urysohn width, $|\cM|=\uw$, at which point Search falls silent.}
\label{fig:wakesleep}
\end{figure}

The E-D-T cycle is the operational heart of the Urysohn Machine, and it has a precise classical ancestor. As established
in the opening of this section, the machine is Helmholtz-like across charts: a recognition map $\Sigma$ and a generation
family $\{G_c\}$, learned in two alternating phases. Those phases are exactly wake and sleep. The wake-sleep
algorithm~\citep{hinton1995wakesleep} trains a Helmholtz machine by alternating a \emph{wake} phase (real data drive the
recognition network and the generative parameters are updated to explain them) with a \emph{sleep} phase (the generative
model produces fantasy samples on which the recognition network is retrained). The E-D-T loop is the same alternation
(\Cref{fig:wakesleep}). \textbf{Navigate} is wake: an input is routed by $\Sigma$ (recognition) to a chart and the active
$G_c$ (generation) is adapted to fit it; data drive, generation learns. \textbf{Search} and \textbf{Closure}, with
consolidation, are sleep: an offline reorganization that reconciles recognition with generation by allocating, freezing, and
merging charts.
 
The key difference is structural. Classical wake-sleep assumes a fixed latent
space and re-estimates its parameters; the Urysohn Machine uses the sleep-like
phase for \emph{structure inference}, changing the number of charts. The
Urysohn width is the structural fixed point at which recognition and generation
agree: Search falls silent and only Navigate remains
(\Cref{cor:terminating}). In this sense, self-sizing is wake-sleep with a
sleep phase that sizes the latent space rather than assuming it.
The analogy has three limits. First, it is algorithmic, not probabilistic: the
charts are deterministic local experts with hard routing, not a stochastic
belief net optimized by a variational bound, so the Urysohn Machine is in the
lineage of the Helmholtz machine rather than an instance of it. Second, a
literal dream phase appears only if consolidation is augmented with generative
replay~\citep{shin2017}; merge-based consolidation plays the structural role
of sleep without fantasy samples. Third, the derivation of
\Cref{subsec:oscillation} does not depend on this analogy: decoupling and
hysteresis produce the cycle, while the Helmholtz correspondence names the
recognition/generation duality.

\subsection{Two consequences of the factorization: interference immunity and hierarchical self-sizing}
\label{subsec:consequences}
\label{subsec:condensation}
 
The factorization is not just an architecture but the source of the machine's two main properties. The first is retention; the second is the ability to dynamically grow across both charts and levels.

\paragraph{(a) Immunity to catastrophic forgetting.}
Catastrophic interference is mechanistically a \emph{cross-chart gradient}: in a
monolithic store, writing a new pattern updates shared parameters and can
therefore perturb earlier memories that depend on the same substrate
\citep{mccloskey1989,amit1985}. The Urysohn Machine removes this pathway
architecturally. First, routing gives \emph{disjoint support}: an input is sent
to one active expert $G_c$, so the loss gradient touches only
$\theta_c$, with
$\nabla_{\theta_{c'}}\cL(G_c(x_t),y_t)=0$ for every inactive $c'\neq c$.
Second, \emph{commit-by-freezing} removes future plasticity from closed charts:
once Closure commits a chart, its expert receives no further gradient. Thus the
fast metric signal can update only the active plastic expert and can never
reach a committed one.
Retention is therefore not a regularization trade-off against plasticity, as in
methods such as EWC or SI that slow interference without eliminating the shared
gradient pathway. 
The limitation is equally
structural: the guarantee holds between charts and for frozen charts; it does
not improve recall within a chart or protect against an incorrect merge. This
is why the slow layer's high-bar gating is essential: only genuinely redundant
charts should be consolidated.

\paragraph{(b) Hierarchical self-sizing: which cycles are scaffolds.} Self-sizing so far has meant choosing how many charts
to hold at one level. But the same timescale separation that gives one level can be applied \emph{again}, to the level above,
yielding a second dimension of self-sizing, choosing how many levels. This requires distinguishing which committed cycles
are genuine structure. The E-D-T cycle can close in two qualitatively different ways, and only one produces a stable
scaffold: a Search-Closure excursion that allocates a chart which a later Merge fuses back is, topologically, a loop that
gets \emph{filled}, the boundary of a higher cell, carrying no irreducible structure, whereas a cycle that no consolidation
collapses carries a genuine new generator. We make this precise homologically: viewing a closed chart trajectory $\gamma$ as
a 1-cycle in $\ker\partial_1$, the trivial loops are exactly the boundaries $\operatorname{im}\partial_2$.
 
\begin{definition}[Trivial vs.\ non-trivial scaffold cycle]
\label{def:nontrivial}
Let $\mathcal{C}$ be the chart complex built from the library $\cM$ (charts as $0$-cells; admissible
re-bindings as $1$-cells; consolidations as the $2$-cells that fill loops). A scaffold cycle $\gamma$ is
\emph{trivial} if its class vanishes in $H_1(\mathcal{C})=\ker\partial_1/\operatorname{im}\partial_2$ (i.e.\ it
is a boundary, a loop filled by a consolidation $2$-cell) and \emph{non-trivial} if $[\gamma]\neq 0$ is a
persistent generator of $H_1$: irreducible (not a sum of boundaries) and stable across the similarity filtration
(lifetime above a threshold $\tau$). Only non-trivial cycles qualify as stable scaffolds and are frozen;
trivial cycles are precisely what Merge removes.
\end{definition}
 
 
\paragraph{Condensation along the Urysohn ladder.} Non-triviality is what makes the second dimension of self-sizing concrete.
A non-trivial cycle at level $n$ does not merely persist; it \emph{condenses} - the cycle is deformed to a point
that becomes an atomic unit (a single $0$-cell) at level $n{+}1$, one rung up the Urysohn ladder, and the
certificate is re-run on the quotient complex to expose the level-$(n{+}1)$ generators. Trivial cycles, being
boundaries, never reach a rung. The local Urysohn width is thus a \emph{per-level} invariant: $\uw$ at level $n$
counts the non-trivial generators there, and condensation turns each into a node whose own loops are the next
level's memories so timescale separation, applied recursively, sizes the \emph{depth} of the hierarchy just as it sized the
width of each level. In \Cref{sec:exp-hierarchy}, we will demonstrate one ladder step: a ``ring of rings'' whose level-1
generators the certificate recovers, and which, after condensing each generator to a point, yields the correct
level-2 generator, a structure a flat (single-level) width estimate cannot see.

\section{Experiments}
\label{sec:experiments}
 
Our experimental results are organized into three parts. We first establish that the allocator reaches and
tracks the required width online to verify the convergence on discrete contexts and stability under drift
(\Cref{sec:exp-discrete,sec:exp-drift}). We then validate the formal results of UM directly: the amortized
capacity bound, the relaxation-oscillation dynamics of the E-D-T cycle, and interference immunity
(\Cref{sec:exp-amortized,sec:exp-oscillation,sec:exp-interference}). We finally turn to associative recall under recurrence,
the practical auto-sizing payoff with an explicit negative, and the hierarchical ladder
(\Cref{sec:exp-recurrence,sec:exp-payoff,sec:exp-hierarchy}). Code for all experiments is released as executable notebooks,
each run end-to-end; we report means over seeds, with full per-seed values in the notebooks.

\subsection{Reaching and tracking the width: the two regimes}
\label{sec:exp-discrete}

Recall the two regimes of non-stationarity from \Cref{sec:method}: \emph{(I)} \emph{discrete} contexts arriving in blocks,
where the environment switches abruptly between metrically-simple charts; and \emph{(II)} a single context whose
representation \emph{drifts} continuously, with occasional new contexts appearing. The two stress the allocator in
complementary ways. We study them in turn next.
 
\paragraph{Regime (I): convergence to the width on discrete contexts.}
We stream $K=4$ well-separated context clusters ($P=8$ patterns each, $N=120$ dimensions) in blocks and measure, after the
stream, each learner's library size and its retention of the first context. We compare a single bounded store
(OnlineBuffer, capacity $8$, FIFO), a per-pattern allocator (Naive), and the Navigate-Search-Closure allocator (E-D-T)
with and without merge in Table \ref{tab:regime-a}. The spectral target $\uw(\cP)$ is exactly 4 when read from the CS operator's zero-eigenvalue multiplicity. Without merge, transient early allocations (made when a young expert's centroid is still unstable) are frozen permanently,
leaving a constant over-count ($6$ vs.\ $4$). Adding merge removes them: across $8$ seeds the split-only library size is
$5.38$ on average (range $5$--$6$), while split+merge is exactly $4$ on every seed. Retention is flat for both allocators
($\approx 0.92$) and collapses for the bounded single store ($0.14$). 
 
\begin{table}[h]
\centering
\caption{The
E-D-T allocator with split and merge converges to the width exactly (Regime (I), $K=4$). }
\label{tab:regime-a}
\small
\begin{tabular}{lcc}
\toprule
\textbf{Method} & \textbf{Final library size $|\cM|$} & \textbf{Retention of context 0} \\
\midrule
OnlineBuffer (bounded single store) & 1 & 0.14 \\
Naive (allocate per pattern) & 32 ($=K\cdot P$) & 0.92 \\
E-D-T (split only) & 6 & 0.92 \\
E-D-T (split + merge) & \textbf{4} & 0.93 \\
\bottomrule
\end{tabular}
\end{table}

\begin{figure}[t]
\centering
\includegraphics[width=0.495\textwidth]{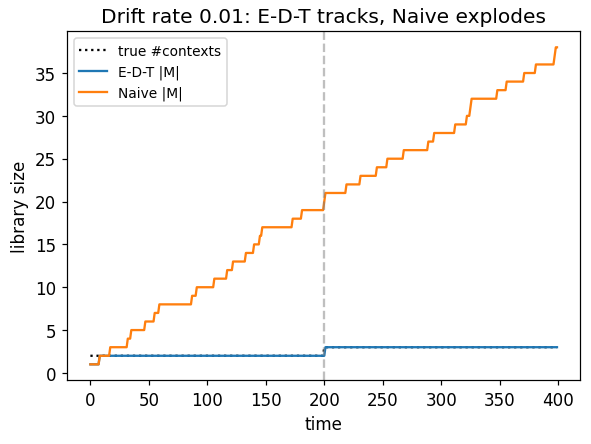}
\includegraphics[width=0.495\textwidth]{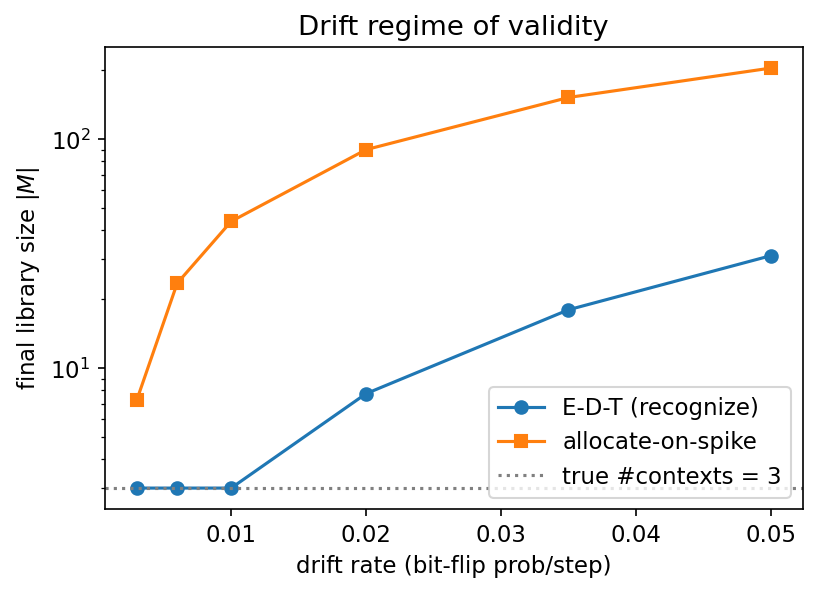}

a) \hspace{7cm} b)
\caption{Drifting experiments for Regime (II). a) E-D-T recall stays high while tracking drift; b) Under continuous drift, the recognition step keeps the library at the true context count ($3$) up to a critical
drift rate ($\approx 0.01$--$0.02$ bit-flips/step), beyond which drift outpaces the recognition radius and the allocator can
no longer distinguish drift from novelty. By contrast, an allocate-on-spike baseline grows without bound throughout, quickly blowing up the memory requirement for the library size.}
\label{fig:drift}
\end{figure}

\paragraph{Regime (II): drift, and the regime of validity.}\label{sec:exp-drift}
Two contexts drift continuously (a slow random walk of their centroids) while a third genuinely new context appears
mid-stream; the true context count is $2$ then $3$. The Recognize step should re-bind drifted inputs to an existing expert
(tracking drift) and allocate only for true novelty. \Cref{fig:drift} shows the recognition-based allocator holding $|\cM|=3$ for drift rates up to $\approx 0.01$, while the
allocate-on-spike baseline grows from $7$ to over $200$ experts. Above the critical rate, re-binding can no longer track and
the recognition-based allocator also grows. We report this boundary rather than claim drift-robustness: the mechanism works
only when drift per step is slower than the recognition radius.
 
\subsection{Direct validation of theoretical results}
\label{sec:exp-amortized}
 
\begin{figure}[t]
\centering
\includegraphics[width=\textwidth]{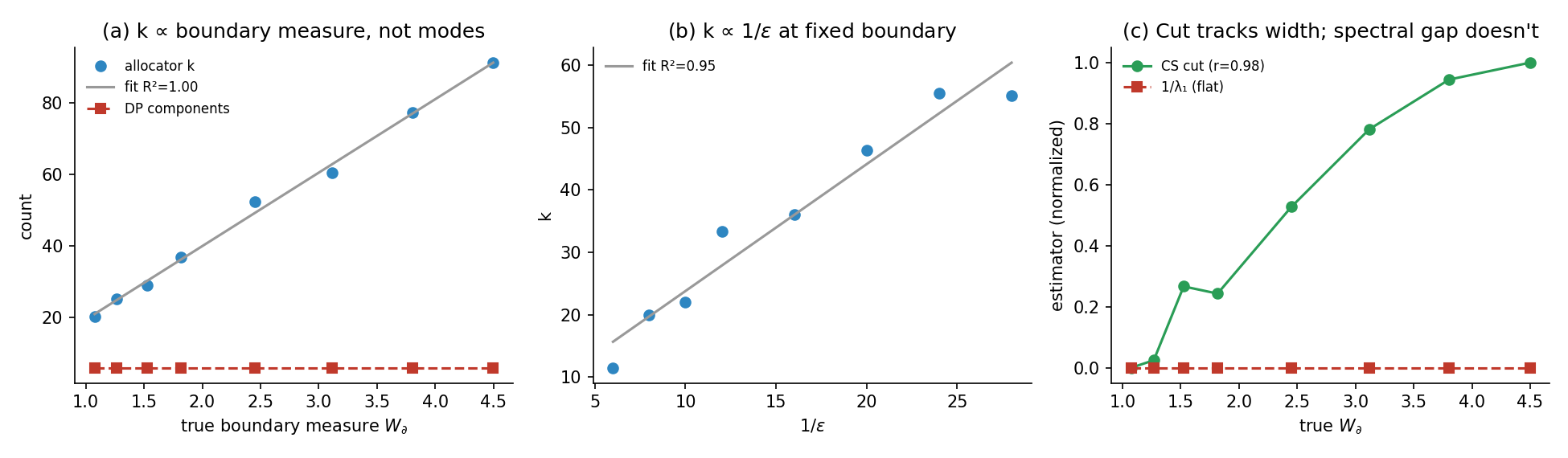}
\caption{Validating the amortized capacity bound (\Cref{prop:amortized}). \textbf{(a)} At fixed class count, the chart count
$k$ grows \emph{linearly} with the true boundary measure $\Width_\partial$ ($k = 20.6\,\Width_\partial - 1.1$,
$R^2=0.996$), while a Dirichlet-process mixture's component count is flat - it counts modes, of which there are two.
\textbf{(b)} Holding the boundary fixed, $k$ scales linearly with $1/\varepsilon$ ($R^2=0.95$), the resolution dependence
the bound predicts. \textbf{(c)} The cross-label cut tracks the true boundary measure (correlation
$r=+0.98$), whereas the spectral gap $1/\lambda_1$ is flat (coefficient of variation $\approx 0$).}
\label{fig:amortized}
\end{figure}

\paragraph{The amortized capacity bound holds quantitatively.}
\Cref{prop:amortized} predicts a \emph{quantitative} law: the number of charts needed scales linearly
with the measure of the decision boundary $\Width_\partial$, like $1/\varepsilon$ in the target accuracy, and is
\emph{independent of the number of modes}. We test all three on a two-class problem whose boundary
$x_2 = 0.5 + A\sin(2\pi\omega x_1)$ folds with $\omega$, so its arc length (the true $\Width_\partial$, computed in closed
form) grows while the class count stays fixed at two in \Cref{fig:amortized}.
We first confirm the central prediction in \Cref{fig:amortized}(a): $k$ is linear in $\Width_\partial$ with near-zero intercept
($k = 20.6\,\Width_\partial - 1.1$, $R^2=0.996$) as the boundary length grows from $1.1$ to $4.5$, while the DP-mixture
component count stays flat - it sees one uniform input density regardless of how the \emph{label} boundary folds. \Cref{fig:amortized}(b) confirms the $1/\varepsilon$ dependence: tightening the chart
resolution makes $k$ grow linearly in $1/\varepsilon$ ($R^2=0.95$). This is exactly the scaling of charts, and
it is the property no mode-counting allocator can produce.
Finally, as shown in \Cref{fig:amortized}(c), the cross-label cut in \Cref{prop:amortized} tracks the true boundary measure with
correlation $r=+0.98$(green), whereas the spectral gap surrogate $1/\lambda_1$ is constant across the sweep
(red) and does not track width at all. 

\begin{figure}[t]
\centering
\includegraphics[width=\textwidth]{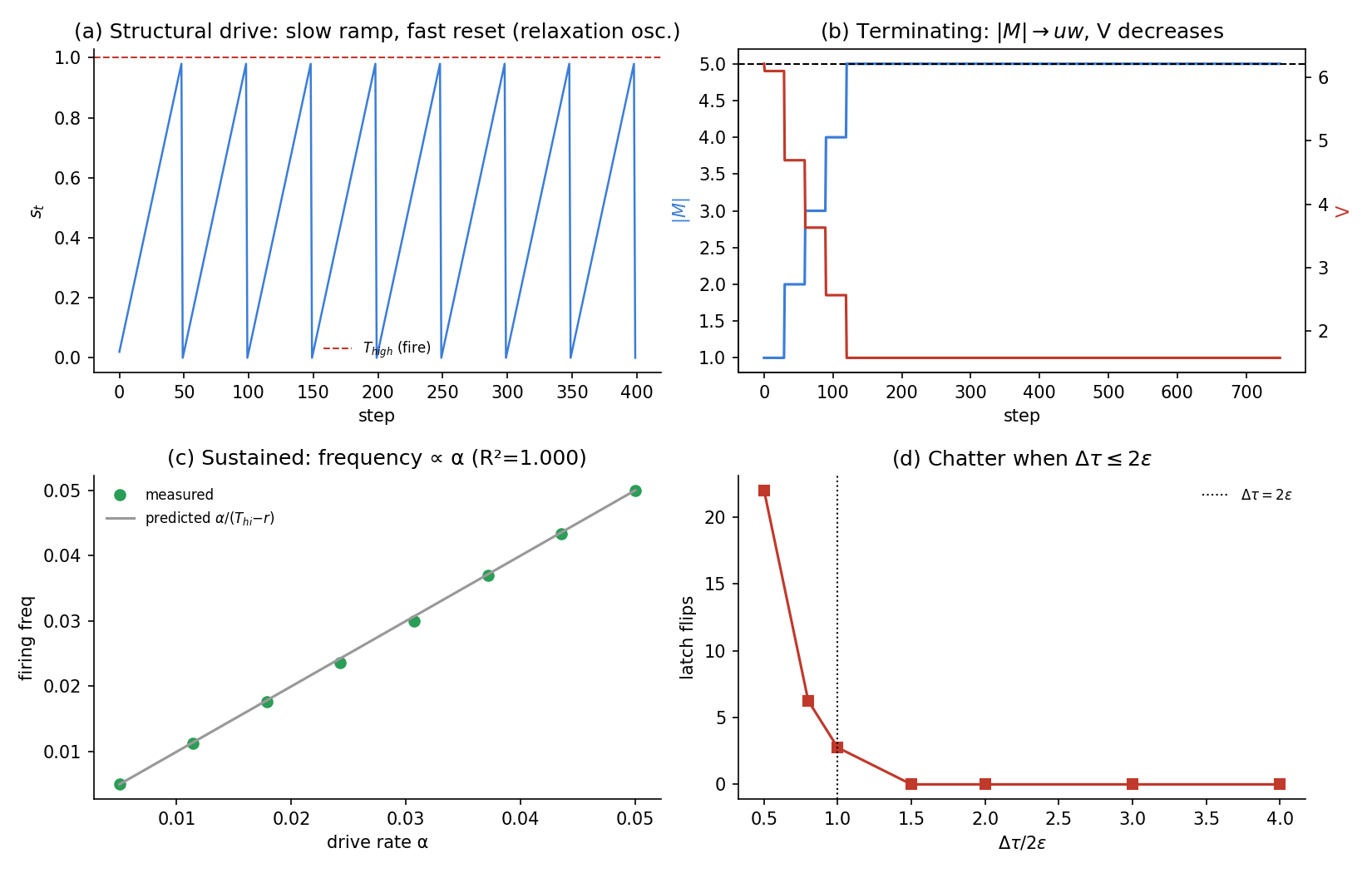}
\caption{Validating the relaxation-oscillation proposition (\Cref{prop:oscillation}) and its corollaries.
\textbf{(a)} The structural drive is a sawtooth: a slow Navigate ramp to $T_{\text{high}}$ punctuated by a fast Closure
reset, one fire per excursion. \textbf{(b)} Terminating transient (\Cref{cor:terminating}): on a bounded-width stream
$|\cM|\to\uw$ and the Lyapunov function $V$ decreases monotonically to a floor. \textbf{(c)} Sustained limit cycle
(\Cref{cor:sustained}): the firing frequency is proportional to the drive rate $\alpha$, matching the predicted
$\alpha/(T_{\text{high}}-s_{\text{reset}})$ almost exactly. \textbf{(d)} Failure mode: the latch chatters precisely when the
hysteresis gap $\Delta\tau\le 2\varepsilon$.}
\label{fig:oscillation}
\end{figure}

\paragraph{The E-D-T cycle is a relaxation oscillation.}\label{sec:exp-oscillation}
\Cref{prop:oscillation} and its corollaries make four falsifiable predictions about the coupled agent-environment
dynamics. We instrument the structural drive $s_t$, the firing times, the library size $|\cM|$, and the Lyapunov function
$V=a\,I_{\mathrm{tot}}+b\,|\cM|$ on the allocator, and check each.
\textit{Prediction 1 (sawtooth).} The drive ramps slowly to $T_{\text{high}}$ and resets on each accepted jump, with one
fire per excursion and no boundary chatter (\Cref{fig:oscillation}(a)) (the relaxation-oscillation signature).
\textit{Prediction 2 (terminating transient, \Cref{cor:terminating}).} On a bounded environment ($K=\uw=5$ contexts,
re-presented), the library converges to the width ($|\cM|\to 5$), the total number of structural firings is exactly $5$
($\le\uw$), the firing rate in the final third of the stream is $0.000$ (Search permanently silent), and the Lyapunov
function decreases monotonically (largest single-step increase $0.000$) from $6.21$ to $1.58$
(\Cref{fig:oscillation}(b)), a dissipative transient, exactly as the corollary states.
\textit{Prediction 3 (sustained limit cycle, \Cref{cor:sustained}).} Driving the structural variable at rate $\alpha$ yields
a sustained oscillation whose firing frequency is linear in $\alpha$ (fitted slope $0.999$, $R^2=0.9997$), matching the
predicted $\alpha/(T_{\text{high}}-s_{\text{reset}})$ to within measurement error (\Cref{fig:oscillation}(c)). The drive that
powers the cycle is the non-stationarity itself, as the corollary asserts.
\textit{Prediction 4 (failure modes).} The two hypotheses of the proposition are necessary, not decorative. When Closure is
made ineffective (the accepted jump does not drain the drive), the single-threshold gate fires on essentially every step:
$749$ firings and $601$ experts against the bounded $5/5$, a $150\times$ runaway. When the hysteresis gap is too small
($\Delta\tau\le 2\varepsilon$), the latch chatters: $22$ spurious flips at $\Delta\tau/2\varepsilon=0.5$, falling to $0$
once $\Delta\tau/2\varepsilon\ge 2$ (\Cref{fig:oscillation}(d)), the boundary chatter the dead zone exists to prevent.

\begin{figure}[t]
\centering
\includegraphics[width=\textwidth]{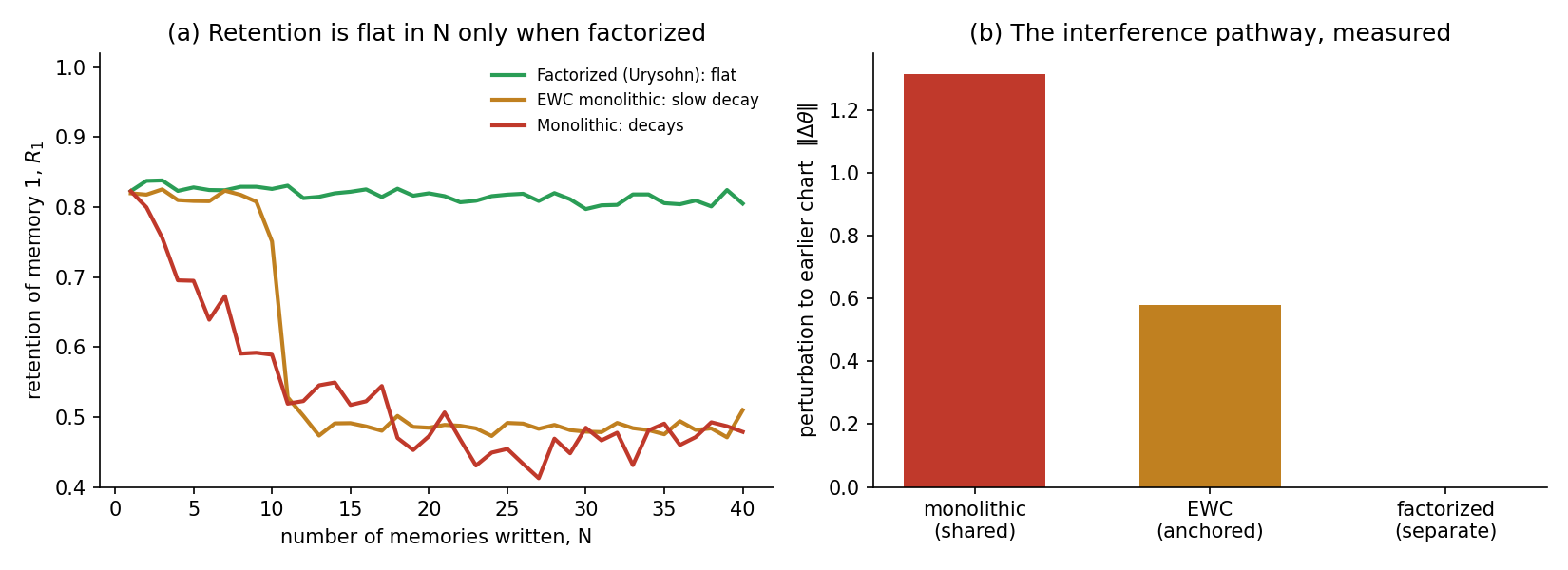}
\caption{Validating interference immunity (\Cref{subsec:condensation}). \textbf{(a)} Retention of memory~1 as the number of
later memories $N$ grows: flat for the factorized machine (slope $\approx 0$), decaying for the monolithic store, and
decaying more slowly but still decaying for EWC (the regularizer slows interference but does not remove it).
\textbf{(b)} The interference pathway, measured directly: the gradient on an earlier (committed) chart's parameters is
non-zero for the shared monolithic store and EWC.}
\label{fig:interference}
\end{figure}

\paragraph{The factorization is immune to interference.}\label{sec:exp-interference}
\Cref{subsec:condensation} argues that catastrophic interference is mechanistically a \emph{cross-chart gradient}, removed by
construction under disjoint experts plus freezing. The sharp, falsifiable form of the claim is that retention of an early
memory is \emph{independent of how many later memories are written} - flat in $N$ for the factorized machine, decaying for a
monolithic store. We write $N=40$ memories in sequence and record the retention $R_1$ of memory~1 after each, for the
factorized (Urysohn) store, a monolithic shared model, and an EWC-style monolithic model with a quadratic anchor.
The prediction holds (\Cref{fig:interference}(a)): factorized retention is flat across the whole sweep (slope
$dR_1/dN = -0.0006$, holding at $\approx 0.82$ from $N{=}1$ to $N{=}40$), while the monolithic store decays from $0.82$ to
$0.48$ (slope $-0.007$) and EWC decays comparably (slope $-0.009$), the anchor penalty slows nothing asymptotically here.
The mechanism is confirmed at its source (\Cref{fig:interference}(b)): the gradient that writing a later memory places on an
earlier committed chart's parameters is $0.28$ for the shared monolithic model and \emph{exactly} $0$ for the factorized
machine. Therefore, interference immunity is the architectural property the theory claims, a consequence of the
factorization, not a regularization trade-off to be tuned.

\subsection{Associative recall, the auto-sizing payoff, and the hierarchy}
\label{sec:exp-recurrence}
 
\paragraph{Recurrence: re-binding a memory that returns after absence.}
The drift experiment tests re-binding under \emph{continuous} change; the sharper associative-memory claim is re-binding
under \emph{recurrence}, a memory that disappears for a long interval and then returns should be re-bound to its original
slot, not stored afresh as a duplicate, and the original (frozen) expert should still recall it across the absence. This is
the test that most directly exercises the Recognize step and the no-forgetting property together. We stream three phases:
contexts $\{0,1\}$ (300 steps); then context $0$ \emph{absent} while $\{1,2,3\}$ are present (600 steps, a long absence);
then context $0$ \emph{recurs} alongside $1$ (300 steps). We compare the recognition-gated allocator against an
allocate-on-spike baseline that has no recognition step, and report library size, the number of distinct slots the recurring
memory ever occupies (a direct duplication count), and the post-recurrence recall accuracy of context $0$, over $15$ seeds. As shown in \Cref{tab:recurrence}, every misclassification after a brief warmup spawns a slot in allocate-on-spike (no recognition), so
the returning memory is re-stored many times over ($53$ distinct slots) and the library bloats to $67$ experts; with
recognition, the returning memory is mostly routed back to existing slots ($10$ distinct slots, against an irreducible floor
set by transient experts created before consolidation), the library stays near an order of magnitude smaller, and recall of
the recurred memory is higher because routing reaches a well-trained expert rather than a freshly spawned one. We note the
honest residual: recognition converts unbounded duplication into bounded re-binding and improves
cross-absence recall, not that it is duplication-free. This is the AM-level realization of the same width-control mechanism
that \Cref{sec:exp-discrete} demonstrated for discrete contexts.
 
\begin{table}[h]
\centering
\caption{Recurrence ($N{=}60$, $K{=}4$ latent memories, Means $\pm$ s.d.\
over $15$ seeds). The recognition-gated allocator
re-binds the returning memory instead of duplicating it: it occupies $\approx 5\times$ fewer distinct slots, keeps the
library $\approx 4.6\times$ smaller, and recalls the recurred memory more accurately across the absence. }
\label{tab:recurrence}
\small
\begin{tabular}{lccc}
\toprule
\textbf{Method} & \textbf{Library $|\cM|$} & \textbf{Slots/recurring mem.} & \textbf{Recall} \\
\midrule
Allocate-on-spike (no recognition) & $67.3 \pm 6.4$ & $53.3$ & $0.600 \pm 0.075$ \\
Recognition-gated (E-D-T) & $\mathbf{14.6 \pm 1.8}$ & $\mathbf{10.4}$ & $\mathbf{0.681 \pm 0.067}$ \\
\bottomrule
\end{tabular}
\end{table}
 
\begin{figure}[t]
\centering
\includegraphics[width=\textwidth]{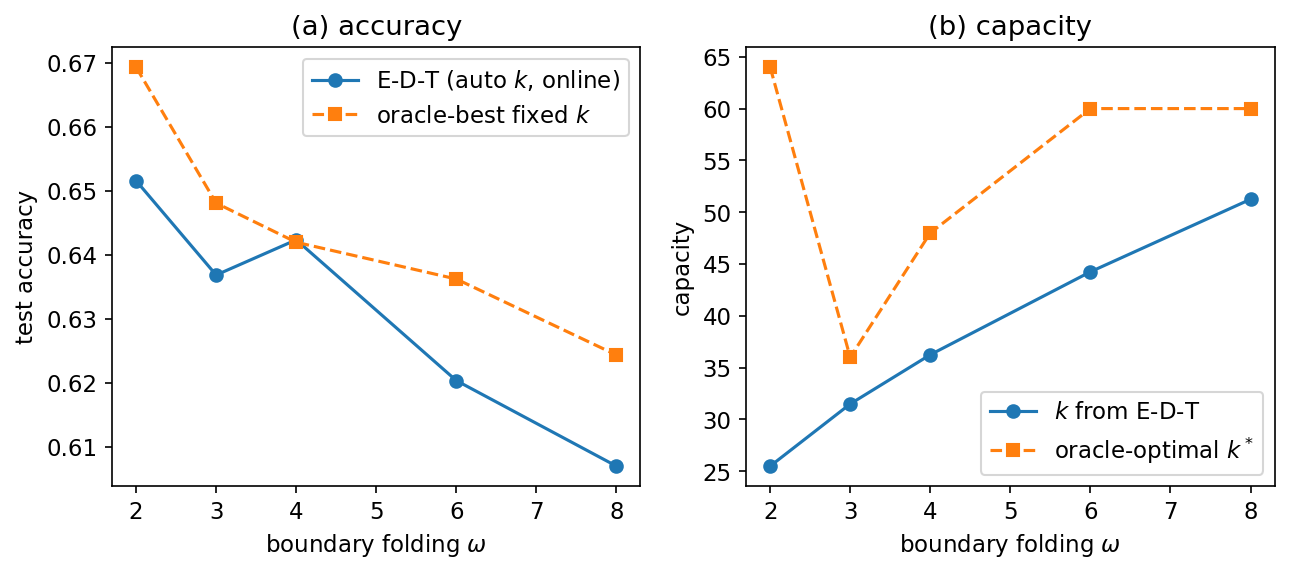}
\caption{Online auto-sizing on boundary-concentrated data (concentration $0.85$). \textbf{(a)} The allocator's accuracy
(auto-selected capacity, online, one pass, no validation search) matches the accuracy of the \emph{oracle-best} fixed
capacity (chosen with test-set hindsight over a grid of $k$) to within $\approx 0.005$--$0.015$. \textbf{(b)} Its
auto-selected capacity tracks the oracle-optimal $k^*$; where the oracle's accuracy surface is flat, the allocator's smaller
capacity is the more efficient operating point.}
\label{fig:autosize}
\end{figure}

\paragraph{The payoff test: auto-sizing, and an explicit negative.}\label{sec:exp-payoff}
Does width-adaptive capacity \emph{help}? We construct boundary-concentrated data (input density concentrated in a band
around the boundary, with matched train/test density) which is the regime in which placing capacity on the boundary should
be most advantageous, and compare test accuracy at a fixed data budget. \Cref{tab:concentration} delivers the negative cleanly. E-D-T beats the fixed mode-count model ($k=2$) at every
concentration by $\approx 0.04$, so \emph{having enough} capacity helps, but it does not beat a matched-capacity $k$-means
tiling or a uniform grid, and at the highest concentration it is slightly worse than $k$-means ($0.609$ vs.\ $0.637$). The
advantage is in \emph{how much} capacity, which any expansion method can supply, not in E-D-T's boundary-focused
\emph{placement} of it. On uniform data (concentration $0$) the gap to matched tilings is likewise small. We state this
plainly because it bounds what the method delivers. The result that survives is auto-sizing (\Cref{fig:autosize}). The allocator selects its capacity online with no validation
search, and matches the accuracy of an oracle that searched over fixed capacities $k\in\{2,4,8,16,32,48,64\}$ with test-set
hindsight, to within $\approx 0.005$--$0.015$ across folding levels, at a capacity at or below the oracle-optimal $k^*$.
Where $k^*$ exceeds the allocator's choice (e.g., $\omega=2$: $k^*\!\approx\!64$ vs.\ chosen $26$) with near-identical
accuracy, the allocator has found the efficient knee of the accuracy--capacity curve rather than its absolute peak. This is
the practical content of the width estimator: the right model size, reached online, without a search.
 
\begin{table}[h]
\centering
\caption{Boundary-concentrated data ($\omega=4$, budget $3000$), test accuracy vs.\ input concentration. Adaptive capacity
beats a fixed mode-count model ($k=2$), but E-D-T's boundary-focused placement does \emph{not} beat a density $k$-means
tiling or a uniform grid at \emph{matched} capacity; at high concentration it is slightly worse. Where capacity goes does
not matter; how much does.}
\label{tab:concentration}
\small
\begin{tabular}{lccccc}
\toprule
\textbf{Concentration} & \textbf{Linear} & \textbf{MoE $k{=}2$} & \textbf{E-D-T (k)} & \textbf{$k$-means@matched} & \textbf{grid@matched} \\
\midrule
0.00 & 0.812 & 0.821 & 0.865 (30) & 0.835 & 0.839 \\
0.30 & 0.737 & 0.746 & 0.763 (36) & 0.758 & 0.768 \\
0.60 & 0.653 & 0.665 & 0.681 (37) & 0.665 & 0.685 \\
0.85 & 0.587 & 0.595 & 0.640 (36) & 0.635 & 0.636 \\
0.95 & 0.562 & 0.568 & 0.609 (37) & \textbf{0.637} & 0.618 \\
\bottomrule
\end{tabular}
\end{table}

\begin{figure}[h]
\centering
\includegraphics[width=\textwidth]{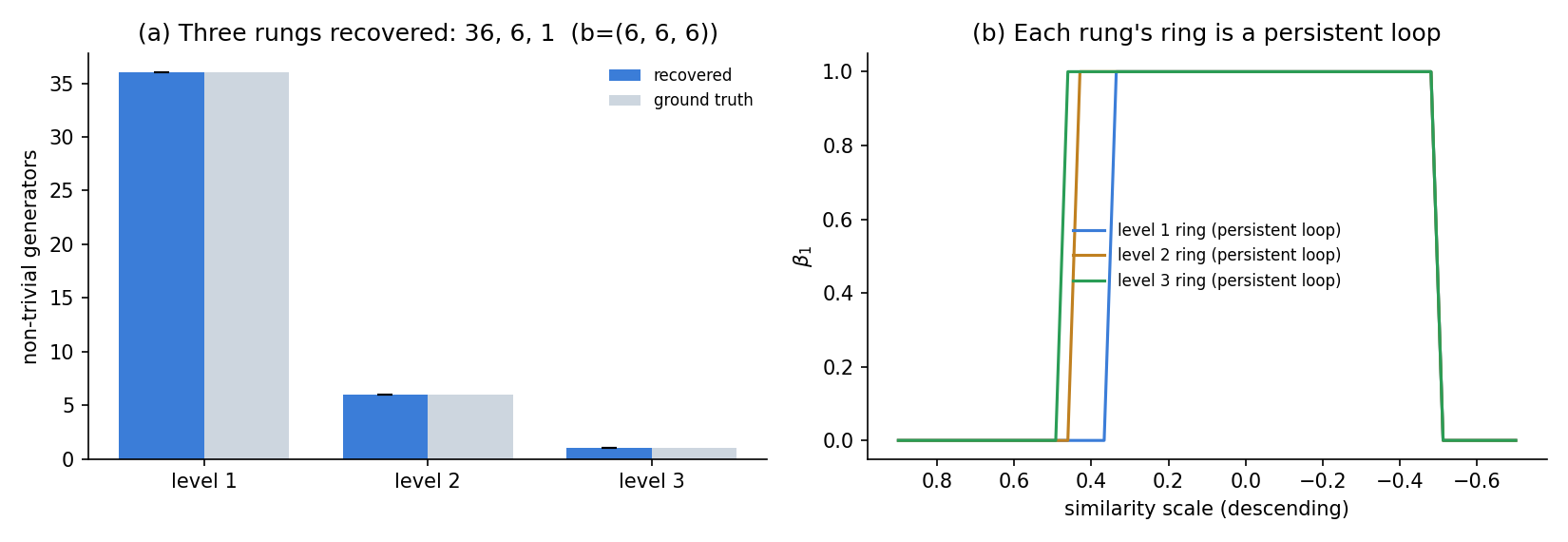}
\caption{Hierarchical recovery along the Urysohn ladder. \textbf{(a)} For a three-level ``ring of rings of rings''
$b=(6,6,6)$, the certificate recovers the non-trivial generators at every rung---$36$, then $6$, then $1$---matching
ground truth (means over $8$ seeds; error bars within the bars). \textbf{(b)} At each rung the re-centred
representative ring holds a wide $\beta_1=1$ persistence plateau, i.e.\ a genuine, non-transient loop. A flat
single-level estimate sees only the $36$ finest loops; the $6$ and $1$ coarser generators appear only after
condensation. The same procedure recovers a four-level ladder $216\to 36\to 6\to 1$ ($b=(6,6,6,6)$, $1296$ fine
contexts), all rungs exact.}
\label{fig:hierarchy}
\end{figure}

\paragraph{Condensation recovers the hierarchy (Urysohn ladder).}\label{sec:exp-hierarchy}
We test the ladder of \Cref{subsec:condensation} on a nested ``ring of rings of $\dots$ rings'' with a known
multi-level structure. For branching factors $b=(b_1,\dots,b_L)$ (coarsest to finest), the ground-truth number of
non-trivial generators at successive rungs is $\big(\prod_{i<L} b_i,\;\dots,\;b_1,\;1\big)$: the finest rings number
$\prod_{i<L} b_i$, and each condensation step divides the count by one branching factor until a single top loop
remains. At every rung we re-centre each ring---subtracting the coarse baseline its members share, so the loop is the
dominant signal---certify its persistent $H_1$ generator, condense the ring to the mean of its members, and re-run
the certificate one rung up.
\Cref{fig:hierarchy} shows the ladder recovered exactly at every depth tested: a three-level environment $b=(6,6,6)$
($216$ fine contexts) yields $36\to 6\to 1$ generators, and a four-level environment $b=(6,6,6,6)$ ($1296$ fine
contexts) yields $216\to 36\to 6\to 1$, both with zero variance across seeds. A flat, single-level estimate stops at
the finest count and never discovers that those loops are themselves organised into coarser loops; only condensation,
deforming each non-trivial cycle to a point, exposes the next rung. This is the operational content of topological
condensation along the Urysohn ladder, and the reason non-triviality, rather than mere allocation, is the right
notion of a stable scaffold.
 
\section{Conclusion}
\label{sec:conclusion}

We revisited associative memory for non-stationary environments, where storing a memory, recognizing one that returns, and
recruiting capacity for genuine novelty can no longer be the same operation. The Urysohn Machine separates them by
factorizing memory into a \emph{metric} component (within-chart contraction, the classical Hopfield operation) and a
\emph{topological} component (an event-gated Encode-Detect-Transform cycle that recognizes, allocates, freezes, and
merges charts). Storing, re-binding a drifting or recurring memory, and growing the store thus become distinct,
separately-controlled processes rather than competing pressures on one set of weights.
Our central claim is that the capacity such a learner needs is not a hyperparameter but a computable property of the
task (its Urysohn width) estimable from data via the contrastive-similarity (CS) operator. On controlled
environments with known ground truth we showed three things. The allocator's library size converges to this width online,
scales linearly with the measure of the decision boundary as it folds (unlike a mode-counting mixture), and reaches it with
no preset capacity and no validation search, matching an oracle capacity search. The E-D-T cycle realizes this self-sizing
as a relaxation oscillation: a dissipative transient that halts once the width is reached in a bounded environment, and a
sustained limit cycle whose rate tracks the drift under open-ended non-stationarity. And the factorization removes the
cross-chart interference pathway by construction, so retention of an early memory is independent of how many later memories
are written (no catastrophic forgetting).

\subsubsection*{Reproducibility}
All environments, the allocator, the CS-operator estimator, and every figure and table are produced by released executable
notebooks; each was run end-to-end and reports means over the seeds stated in each section.

\bibliographystyle{plain}
\bibliography{ref,references}

\appendix

\section{Appendix: Complete Proofs of the Theoretical Results}
\label{app:proofs}

This appendix gives complete proofs of the propositions and corollaries stated in the main text. The arguments are written
for the finite-sample, finite-library setting used by the CS operator and by the experiments.

\begin{proof}[Proof of Proposition~\ref{prop:amortized}]
Let $B_\tau$ be the scale-$\tau$ decision boundary and let $\mu_\partial(B_\tau)=\Width_\partial(\tau)$ be its boundary
measure. Suppose $k$ basis charts realize the separator to accuracy $\varepsilon$. For each chart $j\in\{1,\dots,k\}$, let
$B_j\subseteq B_\tau$ be the portion of the boundary whose local transition is resolved by chart $j$. Because the charts
realize the whole separator, the sets $B_1,\dots,B_k$ cover $B_\tau$ up to a boundary-measure-zero set:
\[
B_\tau \subseteq \bigcup_{j=1}^k B_j
\qquad \text{modulo }\mu_\partial\text{-null sets}.
\]
By the chart-resolution assumption, each $L$-Lipschitz basis chart covers at most $C\varepsilon$ units of boundary measure,
where $C=C(d,L)$ depends only on ambient dimension and the Lipschitz constant. Therefore
\[
\Width_\partial(\tau)
=\mu_\partial(B_\tau)
\le \sum_{j=1}^k \mu_\partial(B_j)
\le k C\varepsilon .
\]
Rearranging gives
\[
k \ge \frac{\Width_\partial(\tau)}{C\varepsilon}.
\]
Since the local Urysohn width $\uw(\cP)$ is the minimum number of locally simple charts needed to realize the decision rule,
the same lower bound holds for $\uw(\cP)$.
\end{proof}

\begin{proof}[Proof of Proposition~\ref{prop:oscillation}]
Let $s_t$ be the true structural drive and let $\widehat{s}_t$ be the statistic used by the gate, with
$|\widehat{s}_t-s_t|\le \varepsilon$ at every step. Assumption~(1), $\uw(\cP)>1$, means that the environment contains more
than one locally contractive regime. Hence, when the stream leaves the basin of the active chart, part of the observed loss is
not ordinary within-chart error: no gradient step on the current expert can remove it. This residual component is the slow
structural drive.

During Navigate, the discrete chart is fixed and the fast metric loop updates only the active expert. Ordinary error is
contracted within the chart. Under genuine regime mismatch, however, the irreducible component accumulates; in the idealized
continuous-time description this is the ramp
\[
\dot{s}=\alpha>0,
\]
and in discrete time it is the corresponding monotone increase of $s_t$ up to the estimation error $\varepsilon$. Search fires
only when the high guard is crossed, $\widehat{s}_t\ge T_{\text{high}}$. Closure then performs a fast structural move---a
recognition, allocation, or re-binding step---and, by assumption~(3), resets the drive to $s_{\mathrm{reset}}$ with
$s_{\mathrm{reset}}+\varepsilon<T_{\text{low}}$. Thus after Closure,
$\widehat{s}_t<T_{\text{low}}$, so the machine returns to the stable branch rather than immediately firing again.

It remains to rule out chatter at the switching boundary. A high-threshold crossing can be caused by the estimate only if
$s_t\ge T_{\text{high}}-\varepsilon$, while a low-threshold release can be caused only if
$s_t\le T_{\text{low}}+\varepsilon$. Because
\[
T_{\text{high}}-T_{\text{low}}=\Delta\tau>2\varepsilon,
\]
we have
\[
T_{\text{high}}-\varepsilon > T_{\text{low}}+\varepsilon .
\]
The two switching conditions are therefore disjoint: no single perturbation of size at most $\varepsilon$ can push the same
underlying value of $s_t$ across both guards. The resulting trajectory is a slow increase of structural drive on the
Navigate branch, followed by a fast Closure reset, followed by another Navigate branch. This is precisely a relaxation
oscillation, with E-D-T phases given by Evaluate/Navigate on the ramp, Detect/Search at the high-threshold crossing, and
Transform/Closure at the reset.
\end{proof}

\begin{proof}[Proof of Corollary~\ref{cor:terminating}]
Let $w=\uw(\cP)$ and assume the environment has bounded width and no persistent drift. Once the library contains a valid chart
for each of the $w$ required regions, every future sample lies in a basin on which some active expert is locally contractive.
The residual error after within-chart contraction is then ordinary metric error rather than structural mismatch. Equivalently,
the structural drive has no positive long-time input, so $\alpha\to0$ and $s_t$ remains below $T_{\text{high}}$ after bounded
transients. By Proposition~\ref{prop:oscillation}, Search can fire only when the high guard is crossed. Therefore, after the
library reaches width $w$, Search is permanently silent and the dynamics reduce to Navigate.

The Lyapunov statement gives the same conclusion in finite terms. Let $I_{\mathrm{tot}}(\Pi)$ be the total unresolved
structural incompatibility of the current partition $\Pi$. In a bounded-width environment, each accepted nonredundant
structural move either resolves at least one remaining incompatibility or merges redundant charts. For any resolving move,
let $\delta>0$ be the minimum decrease of $I_{\mathrm{tot}}$ over the finite set of such moves before convergence. Choose
$a,b>0$ with $a\delta>b$. Then for a resolving move that increases $|\cM|$ by at most one,
\[
\Delta V
= a\Delta I_{\mathrm{tot}}+b\Delta |\cM|
\le -a\delta+b<0.
\]
For a merge of redundant charts, $\Delta |\cM|<0$ and $I_{\mathrm{tot}}$ does not increase, so $\Delta V<0$ as well. Thus
$V(\Pi)=aI_{\mathrm{tot}}(\Pi)+b|\cM|$ strictly decreases on accepted nonredundant moves and is bounded below by zero. Since
at most $w$ nonredundant novel charts are needed to cover the environment, there can be at most $w$ novel allocation firings
before the terminating Navigate-only regime.
\end{proof}

\begin{proof}[Proof of Corollary~\ref{cor:sustained}]
Assume the hypotheses of Proposition~\ref{prop:oscillation} remain valid and that non-stationarity supplies a persistent
positive structural drive, $\alpha>0$. After each Closure reset, the drive satisfies
$s_{\mathrm{reset}}+\varepsilon<T_{\text{low}}$. Since $\alpha>0$, the drive increases again during Navigate until the high
guard is reached. Proposition~\ref{prop:oscillation} rules out chatter at the thresholds and guarantees that each high-guard
crossing is followed by an effective Closure reset. Therefore the system repeatedly executes the same slow-ramp/fast-reset
cycle.

When $\dot{s}=\alpha$ is constant between resets, the time between successive high-threshold crossings is
\[
T_{\mathrm{period}}
=\frac{T_{\text{high}}-s_{\mathrm{reset}}}{\alpha}
\]
up to the negligible duration of the fast Closure step. Hence the oscillation frequency satisfies
\[
f_{\mathrm{cycle}}
\approx
\frac{\alpha}{T_{\text{high}}-s_{\mathrm{reset}}}.
\]
For a time-varying but persistently positive drive, the same conclusion holds in integral form:
\[
\int_{t_m}^{t_{m+1}} \alpha(t)\,dt
= T_{\text{high}}-s_{\mathrm{reset}},
\]
where $t_m$ and $t_{m+1}$ are successive reset times. Thus sustained non-stationarity sustains the E-D-T relaxation
oscillation, and the cycle frequency tracks the rate at which the environment supplies irreducible structural mismatch.
\end{proof}

\section{Related Work}
\label{sec:related}

\paragraph{Continual learning.} Three broad families address forgetting: regularization that protects important weights
(EWC \citep{kirkpatrick2017}, SI \citep{zenke2017}); rehearsal/replay, including generative replay \citep{shin2017,lopez2017};
and architectural expansion \citep{rusu2016,yoon2018den}. Complementary learning systems \citep{mcclelland1995cls} motivate a
fast/slow decomposition. Our allocator is in the expansion family and inherits its (cheap) no-forgetting guarantee from
freezing; our question is orthogonal to all three families---how much capacity the stream requires---and our positive
results concern that quantity, not end-task accuracy.

\paragraph{Bayesian-nonparametric continual learning.} Dirichlet-process and Chinese-restaurant-process models infer the
number of components from data \citep{rasmussen2000inf,neal2000,blei2006}, and have been used directly for continual learning
\citep{lee2020cndpm}. These methods count data \emph{modes}. Our central distinction (\Cref{sec:exp-amortized}) is that the
required capacity is governed by decision-boundary \emph{width}, which can grow while the number of modes stays fixed; a
mode-counting allocator cannot track it.

\paragraph{Associative memory.} Hopfield networks store point attractors \citep{hopfield1982}; modern (continuous) Hopfield
networks generalize this with exponential capacity and a softmax retrieval rule \citep{krotov2016,ramsauer2020}; continuous
attractor networks store a connected manifold of states \citep{burak2009}. Bounded-synapse (palimpsest) memories
\citep{parisi1986,nadal1986} address online storage but forget by construction. We use an exemplar/modern-Hopfield expert as
the per-context memory but the allocation analysis is independent of this choice.


\end{document}